\documentclass[10pt,twocolumn,letterpaper]{article}

\usepackage[pagenumbers]{cvpr} % To force page numbers, e.g. for an arXiv version

\usepackage{graphicx}
\usepackage{amsmath}
\usepackage{amssymb}
\usepackage{booktabs}
\usepackage{tabularx}
\newcolumntype{L}[1]{>{\raggedright\arraybackslash}p{#1}}
\newcolumntype{C}[1]{>{\centering\arraybackslash}p{#1}}
\newcolumntype{R}[1]{>{\raggedleft\arraybackslash}p{#1}}
\DeclareUrlCommand\url{\color{magenta}}
\DeclareUrlCommand\nolinkurl{\color{magenta}}

\usepackage[pagebackref,breaklinks,colorlinks]{hyperref}

\usepackage[capitalize]{cleveref}
\crefname{section}{Sec.}{Secs.}
\Crefname{section}{Section}{Sections}
\Crefname{table}{Table}{Tables}
\crefname{table}{Tab.}{Tabs.}

\def\cvprPaperID{5} % *** Enter the CVPR Paper ID here
\def\confName{Ego4D}
\def\confYear{2022}

\begin{document}

%%%%%%%%% TITLE - PLEASE UPDATE
\title{Intel Labs at Ego4D Challenge 2022:\\A Better Baseline for Audio-Visual Diarization}

%\author{Kyle Min (Team: Intel AI) \\
\author{Kyle Min \\\\
Intel Labs\\
{\tt\small kyle.min@intel.com}
}
\maketitle

%%%%%%%%% ABSTRACT
\begin{abstract}
This report describes our approach for the Audio-Visual Diarization (AVD) task of the Ego4D Challenge 2022. Specifically, we present multiple technical improvements over the official baselines. First, we improve the detection performance of the camera wearer's voice activity by modifying the training scheme of its model. Second, we discover that an off-the-shelf voice activity detection model can effectively remove false positives when it is applied solely to the camera wearer's voice activities. Lastly, we show that better active speaker detection leads to a better AVD outcome. Our final method obtains 65.9\% DER on the test set of Ego4D, which significantly outperforms all the baselines. Our submission achieved 1st place in the Ego4D Challenge 2022.
\end{abstract}

%%%%%%%%% BODY TEXT
\section{Introduction}
\label{sec:intro}
Audio-Visual Diarization (AVD) is a multimodal task where the goal is to identify ``who speaks when'' given a video: More specifically, it is required to detect active speakers and also find the start and end times of speech activities for each speaker. AVD has many real-world applications such as transcription~\cite{yoshioka2019advances,afouras2018deep}, video summarization~\cite{bredin2012segmentation}, and human-robot interaction~\cite{gebru2018audio,stefanov2016look}.

Most of the previous state-of-the-art approaches~\cite{cabanas2018multimodal,chung2019said,kang2020multimodal,xu2021ava} assume that active speakers are always visible in the scene. However, this assumption does not hold for egocentric videos because a camera wearer (CW) is usually not visible. Moreover, egocentric videos have a high range of motion blurs and noise due to the CW's movements, which makes it harder to identify all speakers and their speech activities.

\begin{figure}[t!]
\centering
\includegraphics[width=\linewidth]{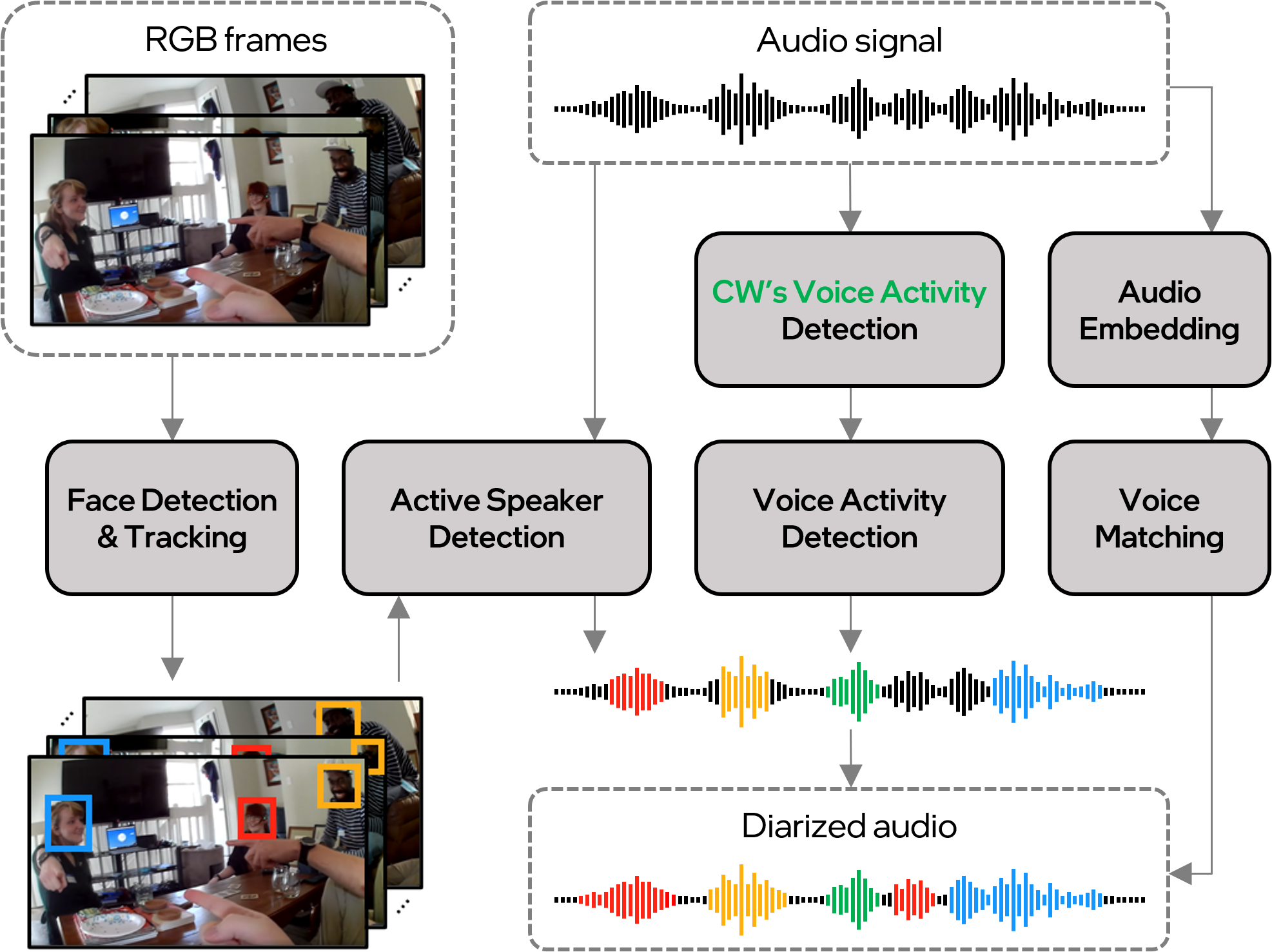}
  \caption{An illustration of our overall framework. First, face regions are detected from the visual signal and connected over time. Second, ASD is performed on the audio-visual input with the detected face regions. Third, we obtain audio embeddings from the audio signal and find potentially missing speech activities. Finally, we detect CW's voice activities and additionally filter out false positives by using an off-the-shelf VAD model. We use four colors (red, blue, yellow, and green) to represent four different speaking identities including CW. Best viewed in color.}
  \label{fig:overview}
\end{figure}

In this report, we describe our approach for the AVD task of the Ego4D Challenge 2022, which encourages the development of audio-visual understanding in egocentric videos using the recently released Ego4D dataset~\cite{grauman2022ego4d}. Figure~\ref{fig:overview} illustrates an overview of our framework. Importantly, we present multiple technical improvements over Ego4D's official baselines. First, we improve the detection performance of CW's voice activity by changing the training scheme of its model. Our modification raises the detection mAP score from 72.0\% to 80.4\%. Second, we demonstrate that an off-the-shelf voice activity detection (VAD) model can be used to remove false positives from the CW's voice activities. This discovery boosts the final diarization performance by more than 4\%. Third, we empirically verify that active speaker detection (ASD) plays a huge role in AVD. Specifically, we show that better ASD leads to significantly better AVD on the validation set of the Ego4D dataset, which has not been demonstrated by the official baselines. Our final method obtains 65.9\% DER on the test set, improved from 73.3\% which is acquired by the best baseline. Our submission achieved 1st position in the Ego4D Challenge 2022 leaderboard.

\section{Method}
\label{sec:method}

In this section, we describe our method in detail. As illustrated by Figure~\ref{fig:overview}, the overall flow of our framework mainly consists of six components. Among them, we present our approaches only on CW's VAD, off-the-shelf VAD, and ASD, each of which has meaningful improvements in performance. For other components, we generally follow the baselines used in the original Ego4D paper~\cite{grauman2022ego4d}.

\subsection{VAD for the CW} \label{subsec:cwvad}
The presence of the CW brings many difficulties to AVD in egocentric videos because its face is usually not visible. Therefore, we detect CW's voice activities by only using the audio signal in the following ways: First, for every 5ms, we compute the log-scaled spectrogram of the audio signal in the low-frequency range from 0 to 4000 Hz. Second, at each time step, we slice the spectrogram over a window of 0.32s, which makes a sequence of single-channel images with a resolution of 256$\times$64. Next, we train a ResNet~\cite{he2016deep} on the generated images using the training set of the Ego4D dataset where the detection performance is optimized on the validation set. During inference, we use the optimized ResNet to directly determine whether the CW is speaking or not for each time step.

\subsection{Off-the-shelf VAD} \label{subsec:vad}
VAD is to identify the presence of human speech from the audio signal at any given moment, thus we can leverage an off-the-shelf VAD model as a separate pre- or post-processing component. For example, it can be used to pre-process the audio signal to detect potential speech activities. Then, an ASD model or the ResNet for the CW's VAD would be applied to the potential results. Alternatively, we can utilize it to post-process the results of the ASD or CW's VAD models.

In particular, we use a popular VAD model called Silero~\cite{Silero_VAD} (we refer to it as SilVAD) that is pre-trained on a large-scale dataset in multiple languages. Here, we demonstrate that SilVAD effectively removes the false positives from the CW's speech activities when it is solely used as a post-processing unit for the CW's VAD. Interestingly, SilVAD does not provide any additional benefit when it is applied for ASD or pre-processing.

\subsection{ASD} \label{subsec:asd}
We significantly improve the final AVD performance by using a better ASD model. Specifically, we use \href{https://github.com/kylemin/SPELL}{\textbf{SPELL}}~\cite{min2022learning}, which is a state-of-the-art ASD method where its effectiveness is validated on AVA-ActiveSpeaker dataset~\cite{roth2020ava,minintel}. We show that improved ASD performance directly leads to better AVD results on the validation set of the Ego4D dataset, which has not been observed in the Ego4D paper.

\section{Experiments}
\label{sec:exp}

\subsection{Implementation details} \label{subsec:imp}
We use a 2D ResNet-18~\cite{he2016deep} in VAD for the CW. It is trained with a batch size of 256 using the Adam optimizer~\cite{kingma2014adam}. In the training process, the dropout is set to 0.5 and the learning rate is fixed at 5$\times10^{-4}$. We apply version 3.1 of SilVAD~\cite{Silero_VAD} as an off-the-shelf VAD. For SPELL~\cite{min2022learning}, we utilize the official code with its default hyperparameters. The whole training process takes less than an hour using a single GPU (TITAN V).

\subsection{Performance of VAD for the CW} \label{subsec:pcwvad}

\begin{table}[t]
\centering
\resizebox{\linewidth}{!}{
\begin{tabular}{L{5.5cm}|C{1.75cm}}
\toprule
\textbf{Method} & \textbf{mAP(\%)}$\uparrow$ \\ \midrule
Energy filtering + Audio Matching~\cite{grauman2022ego4d} & 44.0 \\
Spectrogram + ResNet-18~\cite{he2016deep,grauman2022ego4d} & 72.0 \\
Spectrogram + ResNet-18~\cite{he2016deep} (Ours) & \textbf{80.4} \\
\bottomrule
\end{tabular}
}
\caption{Performance comparison of our method with the baselines used in the Ego4D paper~\cite{grauman2022ego4d} on the validation set of the Ego4D dataset. We report the mAP (mean average precision) of the detected CW's voice activities.}
\label{tab:pcwvad}
\end{table}

We compare our method for the CW's VAD with the baselines in Table~\ref{tab:pcwvad}. The results indicate that our modification is effective in improving the detection performance.

\subsection{Impact of post-processing} \label{subsec:pp}

\begin{table}[t]
\centering
\resizebox{\linewidth}{!}{
\begin{tabular}{L{1.9cm}|C{4.1cm}|C{2cm}}
\toprule
\textbf{ASD Model} & \textbf{Post-processing w/ SilVAD} & \textbf{DER(\%)}$\downarrow$ \\ \midrule
SPELL~\cite{min2022learning} & No & 70.7 \\
SPELL~\cite{min2022learning} & Yes (on ASD) & 71.9 \\
SPELL~\cite{min2022learning} & Yes (on CW's VAD) & \textbf{66.6} \\
\bottomrule
\end{tabular}
}
\caption{Performance of our method with different post-processing conditions on the validation set of the Ego4D dataset. We report DER (diarization error rate), where a lower value indicates better AVD performance.}
\label{tab:pp}
\end{table}

\begin{table*}[t]
\centering
\resizebox{\linewidth}{!}{
\begin{tabular}{L{2.6cm}|C{3.4cm}|C{3.4cm}|C{3.4cm}}
\toprule
\, \textbf{ASD Model} & \textbf{ASD mAP(\%)}$\uparrow$ & \textbf{ASD mAP@0.5(\%)}$\uparrow$ & \textbf{AVD DER(\%)}$\downarrow$ \\ \midrule
\, RegionCls~\cite{grauman2022ego4d} & - & 24.6 & 80.0 \\
\, TalkNet~\cite{tao2021someone} & - & 50.6 & 79.3 \\
\, SPELL~\cite{min2022learning} & \textbf{71.3} & \textbf{60.7} & \textbf{66.6} \\
\bottomrule
\end{tabular}
} 
\caption{ASD and AVD performance comparisons of our method with the baselines used in the Ego4D paper~\cite{grauman2022ego4d} on the validation set of the Ego4D dataset. \textbf{We report two metrics to evaluate ASD performance: mAP quantifies the ASD results by assuming that the face bound-box detections are the ground truth (i.e. assuming the perfect face detector), whereas mAP@0.5 quantifies the ASD results on the detected face bounding boxes (i.e. a face detection is considered positive only if the IoU between a detected face bounding box and the ground-truth exceeds 0.5).} We compute mAP@0.5 by using the official evaluation tool provided by Ego4D\protect\footnotemark. For SPELL, better ASD performance leads to a significantly better AVD outcome.}
\label{tab:asdavd}
\end{table*}

In Table~\ref{tab:pp}, we show the performance of our method on the validation set of the Ego4D dataset with different post-processing options. We can observe that post-processing boosts the AVD performance only when it is applied to CW's VAD. We believe that this is because the intensity of the CW's speech is usually higher than other people so the CW's voice activities can be detected relatively easily by the off-the-self VAD model. We also think that post-processing on ASD has a negative impact because it is challenging to detect speech activities of people who are distant from the camera without considering the visual information of their faces. In other words, ASD may need an improved post-processing method other than the off-the-shelf VAD.

\footnotetext{\url{https://github.com/EGO4D/audio-visual/tree/main/active-speaker-detection/active_speaker/active_speaker_evaluation}}

\subsection{Better ASD, Better AVD} \label{subsec:asdavd}

We compare the ASD and AVD performances of our method with the baselines on the validation set of the Ego4D dataset in Table~\ref{tab:asdavd}. Although TalkNet significantly outperforms RegionCls on ASD (50.6\% v.s. 24.6\%), interestingly, the performance difference on AVD is negligible (80.0\% v.s. 79.3\%). However, we can observe that our method using SPELL brings significant performance gains on both ASD and AVD. We believe that this is because our technical improvements provide supplementary benefits for ASD.

\subsection{Final AVD performance} \label{subsec:final}

\begin{table}[t]
\centering
\resizebox{\linewidth}{!}{
\begin{tabular}{L{4.4cm}|C{2.8cm}}
\toprule
\textbf{Method} & \textbf{AVD DER(\%)}$\downarrow$ \\ \midrule
Baseline using TalkNet~\cite{tao2021someone} & 73.3 \\
Ours using SPELL~\cite{min2022learning} & \textbf{65.9} \\
\bottomrule
\end{tabular}
}
\caption{Final AVD performance on the test set of the Ego4D dataset. Our method significantly outperforms the baseline.}
\label{tab:final}
\end{table}

We report the final AVD performance of our method in Table~\ref{tab:final}. When compared to the baseline method using TalkNet, our method achieves 7.4\% lower DER on the test set of the Ego4D dataset. We attained 1st place in the Ego4D Challenge 2022.

\section{Conclusion}
\label{sec:concl}

We have presented multiple technical improvements over the official baselines of Ego4D. First, our modified training scheme results in better optimization and makes VAD for the CW more effective. Second, we show that using the off-the-shelf VAD as a post-processing component improves the AVD performance. Finally, our method using a better ASD model outperforms all the baselines.

%%%%%%%%% REFERENCES
{\small
\bibliographystyle{ieee_fullname}
\bibliography{main}
}

\end{document}